\documentclass{article}
\usepackage{ijcai21}

\usepackage{times}

\usepackage{soul}
\usepackage{url}
\usepackage[hidelinks]{hyperref}
\usepackage[utf8]{inputenc}
\usepackage[small]{caption}
\usepackage{graphicx}
\usepackage{amsmath}
\usepackage{booktabs}
\usepackage{hyperref}
\usepackage{url}
\usepackage{amsmath}

\usepackage{times}
\usepackage{soul}
\usepackage{url}
\usepackage[hidelinks]{hyperref}
\usepackage[utf8]{inputenc}
\usepackage{graphicx}
\usepackage{amsmath}
\usepackage{amsthm}
\usepackage{booktabs}
\usepackage{algorithm}
\usepackage{algorithmic}
\usepackage[space]{grffile}
\usepackage{subcaption}
\usepackage{courier}
\usepackage{amsthm}
\usepackage{xcolor}
\usepackage{listings}
\usepackage{epstopdf}
\usepackage{epsfig}
\usepackage{amsmath}
\usepackage{multirow}
\usepackage{bm}
\usepackage{adjustbox}
\usepackage{enumitem}
\usepackage{amssymb}
\usepackage{caption}

\let\oldnl\nl
\newcommand{\nonl}{\renewcommand{\nl}{\let\nl\oldnl}}

\newtheorem{theorem}{Theorem}
\newtheorem{lemma}{Lemma}
\newcommand{\loss}[0]{{\mathcal{L}}}
\newcommand{\nop}[1]{}
\newcommand{\cwy}[1]{#1}
\newcommand{\citep}{\cite}
\newcommand{\citet}{\cite}

\DeclareMathOperator*{\argmin}{arg\,min}

\usepackage{tikz}

\title{Isotonic Data Augmentation for Knowledge Distillation}

\author{
Wanyun Cui$^1$\footnote{Contact Author}\and
Sen Yan$^2$\\
\affiliations
Shanghai University of Finance and Economics\\
\emails
cui.wanyun@sufe.edu.cn,
woodthree333@gmail.com,
}

\begin{document}

\maketitle

\begin{abstract}
Knowledge distillation uses both real hard labels and soft labels predicted by teacher models as supervision.
Intuitively, we expect the soft labels and hard labels to be concordant w.r.t. their orders of probabilities. However, we found {\it critical order violations} between hard labels and soft labels in augmented samples. For example, for an augmented sample $x=0.7*panda+0.3*cat$, we expect the order of meaningful soft labels to be $P_\text{soft}(panda|x)>P_\text{soft}(cat|x)>P_\text{soft}(other|x)$. But real soft labels usually violate the order, e.g. $P_\text{soft}(tiger|x)>P_\text{soft}(panda|x)>P_\text{soft}(cat|x)$. 
We attribute this to the unsatisfactory generalization ability of the teacher, which leads to the prediction error of augmented samples. Empirically, we found the violations are common and injure the knowledge transfer.In this paper, we introduce order restrictions to data augmentation for knowledge distillation, which is denoted as isotonic data augmentation (IDA).
We use isotonic regression (IR) -- a classic technique from statistics -- to eliminate the order violations. We show that IDA can be modeled as a tree-structured IR problem. We thereby adapt the classical IRT-BIN algorithm for optimal solutions with $O(c \log c)$ time complexity, where $c$ is the number of labels. In order to further reduce the time complexity, we also \cwy{propose} a GPU-friendly approximation with linear time complexity.
We have verified on variant datasets and data augmentation techniques that our proposed IDA algorithms effectively increases the accuracy of knowledge distillation by eliminating the rank violations.
\end{abstract}

\section{Introduction}

Data augmentation, as a widely used technology, is also beneficial to knowledge distillation~\cite{das2020empirical}. For example, \cite{wang2020knowledgethrives} use data augmentation to improve the generalization ability of knowledge distillation. \cite{wang2020neural} use Mixup~\cite{zhang2018mixup}, a widely applied data augmentation technique, to improve the efficiency of knowledge distillation. In this paper, we focus on the mixture-based data augmentation (e.g. Mixup and Cutmix~\cite{yun2019cutmix}), arguably one of the most widely used type of augmentation techniques.

\begin{figure*}[!htb]
	\centering
		\includegraphics[scale=.5]{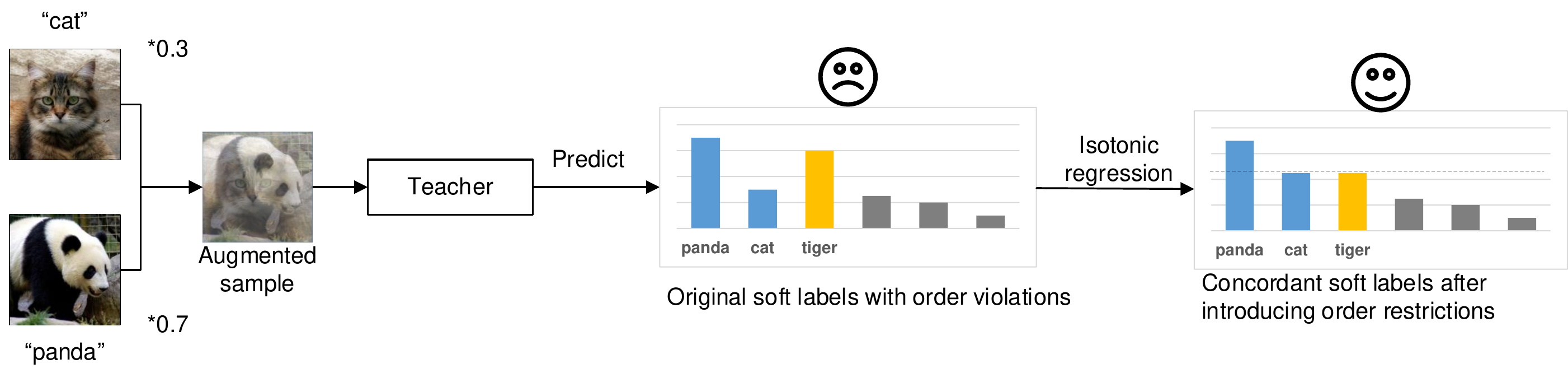}
	\caption{Using isotonic regression to introduce order restrictions to soft labels.}
	\label{fig:intuition_draw}
\end{figure*}

Intuitively, we expect the order concordance between soft labels and hard labels. In Fig.~\ref{fig:intuition_draw}, for an augmented sample $\tilde{x}=0.7*panda+0.3*cat$, the hard label distribution is $P_\text{hard}(panda|\tilde{x})=0.7>P_\text{hard}(cat|x)=0.3>P_\text{hard}(other|\tilde{x})=0$. Then we expect the soft labels to be monotonic w.r.t. the hard labels: $P_\text{soft}(panda|\tilde{x})>P_\text{soft}(cat|\tilde{x})>P_\text{soft}(other|\tilde{x})$. 

\begin{figure}[!tb]
\begin{subfigure}[b]{0.225\textwidth}
	\centering
		\includegraphics[scale=.25]{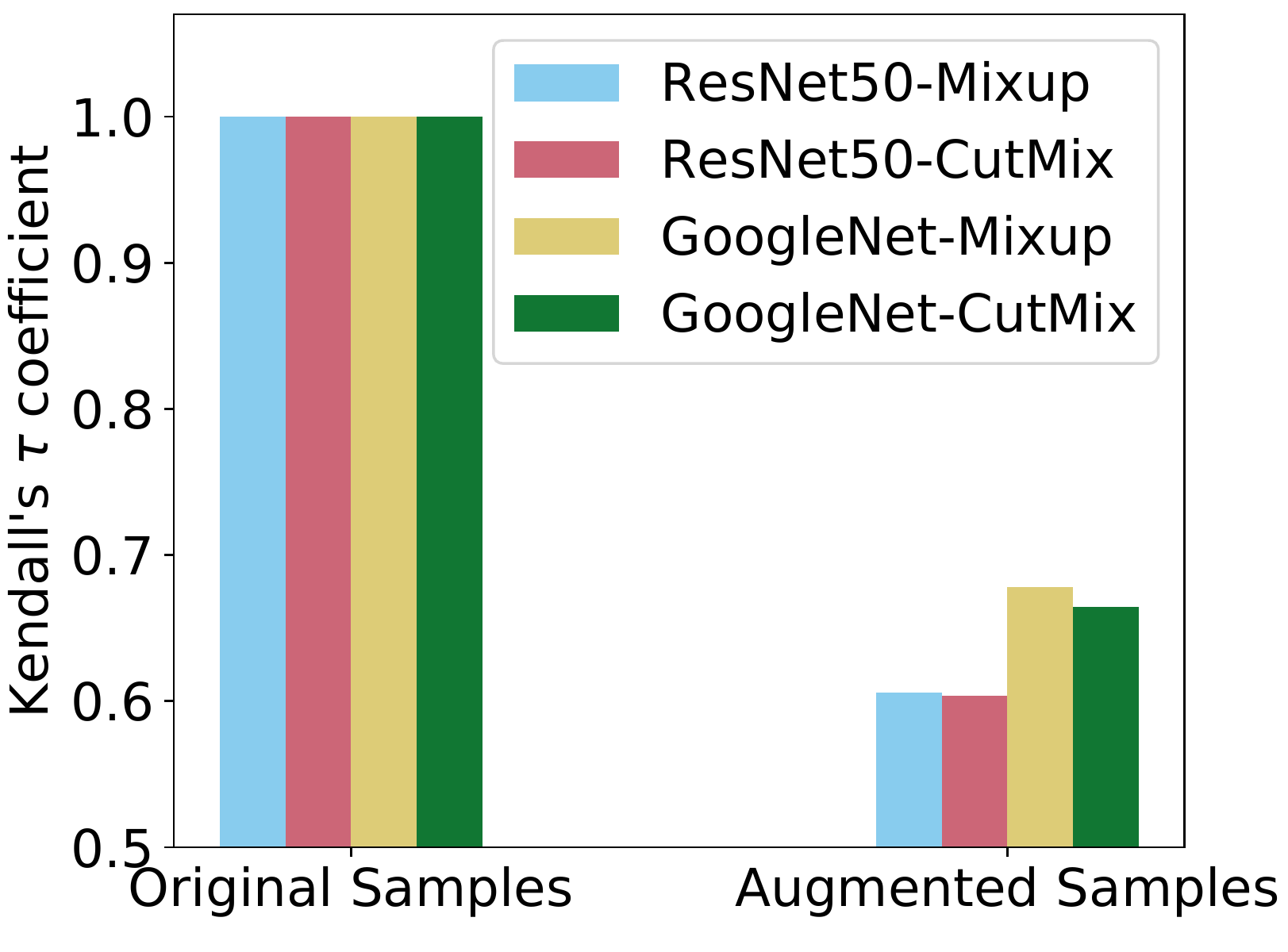}
\caption{The Kendall's $\tau$ coefficient between the soft label distribution and the hard label distribution. Larger $\tau$ means higher ordinal association.}
\label{fig:tau}
\end{subfigure}
\hspace{0.2cm}
\begin{subfigure}[b]{0.225\textwidth}
	\centering
		\includegraphics[scale=.25]{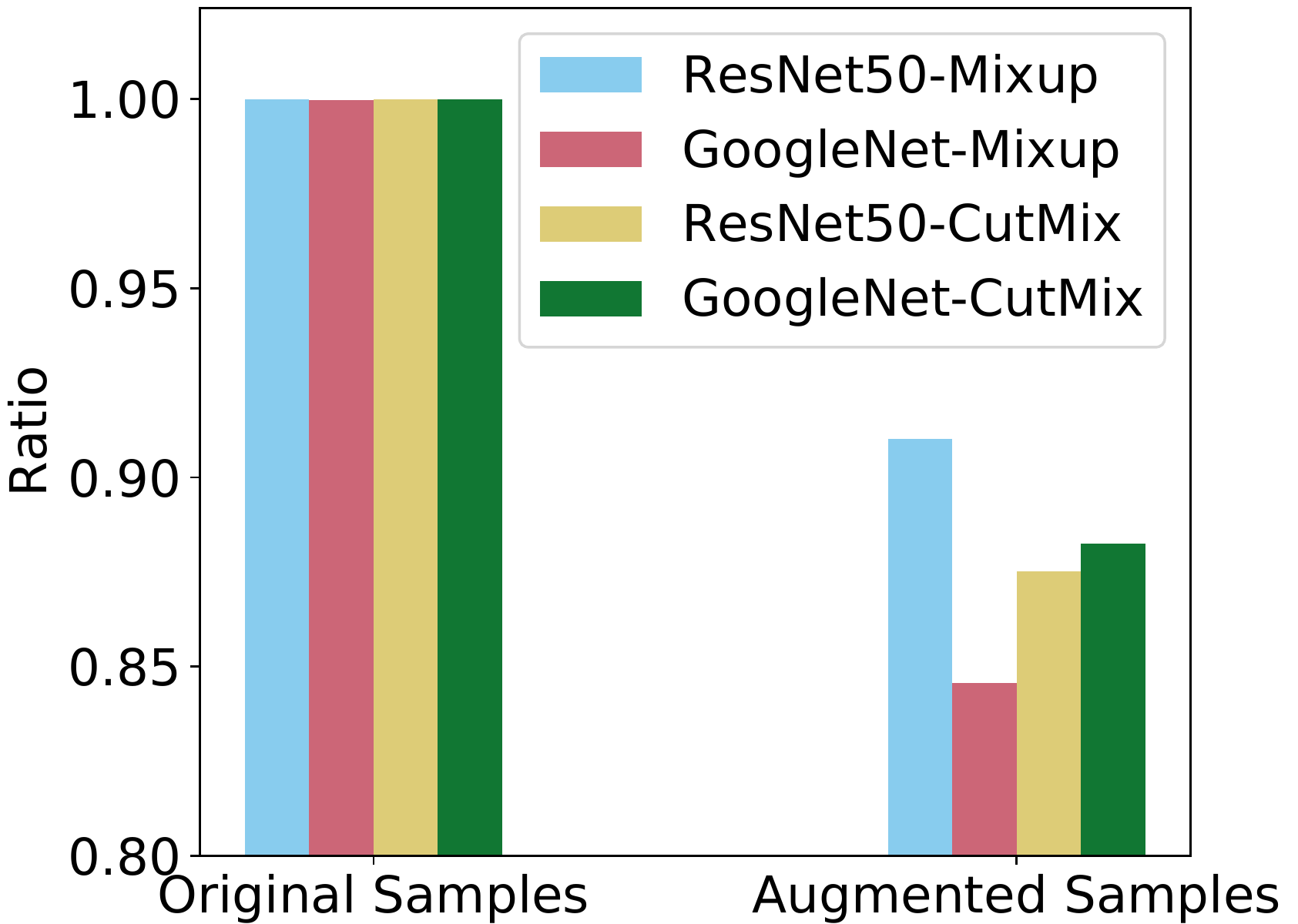}
\caption{The ratio of augmented samples in which at least one original label is in the top $2$ soft labels. \newline}
\label{fig:top2ratio}
\end{subfigure}
	\caption{Both \ref{fig:tau} and \ref{fig:top2ratio} reveal that, the orders of soft labels and hard labels are highly concordant for the original samples. But for augmented samples, the order concordance is broken seriously. 
	This motivates us to introduce the order restrictions in data augmentation for knowledge distillation.}
	\label{fig:effect_adaptive_gamma}
\end{figure}

However, we found critical order violations between hard labels and soft labels in real datasets and teacher models. To verify this, we plot the Kendall's $\tau$ coefficient \cite{10.2307/2332226} between the soft labels and the hard labels of different teacher models and different data augmentation techniques in CIFAR-100 in Fig.~\ref{fig:tau}. We only count label pairs whose orders are known. In other words, we \cwy{ignore} the orders between two ``other'' labels, since we do not know them. A clear phenomenon is that, although the hard labels and soft labels are almost completely concordant for original samples, they are likely to be discordant for augmented samples. What's more surprising is that, in Fig.~\ref{fig:top2ratio}, we \cwy{find} that there are a proportion of augmented samples, in which none of the original labels are in the top 2 of the soft labels.
We attribute this to the insufficient generalization ability of the teacher, which leads to the prediction error of the augmented sample. We will show in Sec~\ref{sec:exp:order} that the order violations will injury the knowledge distillation. As far as we know, the order violations between hard labels and soft labels havn't been studied in previous studies.


A natural direction to tackle the problem is to reduce the order violations in soft labels.
To this end, we leverage the isotonic regression (IR) -- a classic technique from statistics -- to introduce the order restrictions into the soft labels. IR minimizes the distance from given nodes to a set defined by some order constraints. In Fig.~\ref{fig:intuition_draw}, by applying order restrictions to soft labels via IR, we \cwy{obtain} concordant soft labels while keeping the original soft label information as much as possible. IR has numerous important applications in statistics~\cite{barlow1972statistical}, operations research~\cite{maxwell1985establishing}, and signal processing~\cite{acton1998nonlinear}. To our knowledge, we are the first to introduce IR in knowledge distillation. 


Some other studies also noticed the erroneous of soft labels in knowledge distillation and were also working on mitigating it~\cite{wen2019preparing,ding2019adaptive,tian2019contrastive}. However, none of them revealed the order violations of soft labels.

\section{Related Work}
{\bf Knowledge Distillation with Erroneous Soft Labels.} In recent years, knowledge distillation~\cite{hinton2015distilling} as a model compression and knowledge transfer technology has received extensive research interests.
Since the teacher model is non-optimal, how to deal with the errors of soft labels has become an important issue. Traditional methods often solve this problem via optimizing the teacher model or student model.

For teacher optimization, \cite{cho2019efficacy} finds that a larger network is not necessarily a better teacher, because student models may not be able to imitate a large network. They proposed that early-stopping should be used for the teacher, so that large networks can behave more like small networks \cite{mahsereci2017early}, which is easier to imitate. An important idea for teacher model optimization is ``strictness''~\cite{yang2019training}, which refers to tolerating the teacher's probability distribution outside of hard labels. 

The training optimization of the student model is mainly to modify the loss function of distillation. \cite{wen2019preparing} 
proposed to assign different $\tau$s to different samples based on their deceptiveness to teacher models. 
\cite{ding2019adaptive} 
proposed that the label correlation represented by student should be consistent with teacher model. They use residual labels to add this goal to the loss function.

However, none of these studies reveal the phenomenon of rank violations in data augmented knowledge distillation. 

{\bf Data Mixing} is a typical data augmentation method. 
Mixup~\cite{zhang2018mixup} first randomly combines a pair of samples via weighted sum of their data and labels. 
Some recent studies include CutMix~\cite{yun2019cutmix}, and RICAP~\cite{takahashi2019data}.
The main difference among the different mixing methods is how they mix the data. 

The difference between our isotonic data augmentation and the conventional data augmentation is that we focus on relieving the error transfer of soft labels in knowledge distillation by introducing order restrictions. Therefore, we pay attention to the order restrictions of the soft labels, instead of directly using the mixed data as data augmentation. \cwy{We verified in the experiment section that our isotonic data augmentation is more effective than directly using mixed data for knowledge distillation.}


\section{Data Augmentation for Knowledge Distillation}
\subsection{Standard Knowledge Distillation}

In this paper, we consider the knowledge distillation for multi-class classification. We define the teacher model as $\mathcal{T}(x):\mathcal{X} \rightarrow \mathbb{R}^c$, where $\mathcal{X}$ is the feature space, $\mathcal{C}=\{1,\cdots,c\}$ is the label space. We denote the student model as $\mathcal{S}(x):\mathcal{X} \rightarrow \mathbb{R}^c$. The final classification probabilities of the two models are computed by $softmax(\mathcal{T}(x))$ and $softmax(\mathcal{S}(x))$, respectively. We denote the training dataset as $\mathcal{D}_{train}=\{(x^{(1)},y^{(1)}),\cdots,(x^{(n)},y^{(n)})\}$, where $y^{(i)}$ is one-hot encoded. We denote the score of the $j$-th label for $y^{(i)}$ as $y^{(i)}_j$.

The distillation has two objectives: hard loss and soft loss. The hard loss encourages the student model to predict the supervised hard label $y^{(i)}$. The soft loss encourages the student model to perform similarly with the teacher model. We use cross entropy (CE) to measure both similarities:
\begin{equation}
\begin{aligned}
&\loss_\text{hard}(x,y)= CE(softmax(\mathcal{S}(x)),y) \\
&\loss_\text{soft}(x,y)= CE(softmax(\frac{\mathcal{S}(x)}{\tau}),softmax(\frac{\mathcal{T}(x)}{\tau}))
\end{aligned}
\end{equation}
where $\tau$ is a hyper-parameter denoting the temperature of the distillation.

The overall loss of the knowledge distillation is the sum of the hard loss and soft loss:
\begin{equation}
\label{eqn:standardkd}
\small
\begin{aligned}
& \loss_\text{KD} = E_{(x,y) \sim \mathcal{D}_{train}} \alpha \tau^2 \loss_\text{soft}(x,y) + (1-\alpha) \loss_\text{hard}(x,y) \\
\end{aligned}
\end{equation}
where $\alpha$ is a hyper-parameter.

\subsection{Knowledge Distillation with Augmented Samples}
In this subsection, we first formulate data augmentation for knowledge distillation.
We train the student model against the augmented samples instead of the original samples from $\mathcal{D}_{train}$. This method is considered as a baseline without introducing the order restrictions. We then formulate the data augmentation techniques used in this paper.

{\bf Data Augmentation-base Knowledge Distillation.} In this paper, we consider two classic augmentations (i.e., CutMix~\cite{yun2019cutmix} and Mixup~\cite{zhang2018mixup}). Our work can be easily extended to other mixture-based data enhancement operations (e.g. FCut~\cite{harris2020fmix}, Mosiac~\cite{bochkovskiy2020yolov4}). As in Mixup and CutMix, we combine two original samples to form a new augmented sample. For two original samples $(x^{(i)},y^{(i)})$ and $(x^{(j)},y^{(j)})$, data augmentation generates a new sample $(\tilde{x},\tilde{y})$. The knowledge distillation based on augmented samples has the same loss function as in Eq.~\eqref{eqn:standardkd}:
\begin{equation}
\small
\label{eqn:kdaug}
\begin{aligned}
& \loss_\text{KD-aug} = E_{(\tilde{x},\tilde{y}) \sim \mathcal{D}_{train}} \alpha \tau^2 \loss_{soft}(\tilde{x},\tilde{y}) + (1-\alpha) \loss_{hard}(\tilde{x},\tilde{y}) \\
\end{aligned}
\end{equation}
where the augmented sample $(\tilde{x},\tilde{y}) \sim \mathcal{D}_{train}$ is sampled by first randomly selecting $2$ original samples $\{(x^{(i)},y^{(i)}),(x^{(j)},y^{(j)})\}$ from $\mathcal{D}_{train}$, and then mixing the samples.

We formulate the process of augmenting samples as:
\begin{equation}
\begin{aligned}
&\tilde{x}=A(x^{(i)},x^{(j)},\gamma) \\
&\tilde{y}=\gamma y^{(i)} + (1-\gamma) y^{(j)}
\end{aligned}
\end{equation}
where $A$ denotes the specific data augmentation technique. $\tilde{y}$ is distributed in two labels (e.g. $P(panda|\tilde{y})=0.7, P(cat|\tilde{y})=0.3$). We will formulate different data augmentation techniques below.

{\bf CutMix} augments samples by cutting and pasting patches for a pair of original images. For $x^{(i)}$ and $x^{(j)}$, CutMix samples a patch $B=(r_x,r_y,r_w,r_h)$ for both of them. Then CutMix removes the region $B$ in $x^{(i)}$ and fills it with the patch cropped from $B$ of $x^{(j)}$. We formulate CutMix as:
\begin{equation}
A_\text{CutMix}(x^{(i)},x^{(j)},\gamma)=M \odot x^{(i)} + (1-M) \odot x^{(j)}
\end{equation}
where $M \in \{0,1\}^{W\times H}$ indicates whether the coordinates are inside ($0$) or outside ($1$) the patch. We follow the settings in \cite{yun2019cutmix} to uniformly sample $r_x$ and $r_y$ and keep the aspect ratio of $B$ to be proportional to the original image:
\begin{equation}
\begin{aligned}
&r_x \sim Unif(0,W), r_w = W\sqrt{1-\gamma} \\
&r_y \sim Unif(0,H), r_h = W\sqrt{1-\gamma}
\end{aligned}
\end{equation}

{\bf Mixup} augments a pair of sample by a weighted sum of their input features:
\begin{equation}
\label{eqn:mixup}
\begin{aligned}
A_\text{Mixup}=\gamma x^{(i)} + (1-\gamma) x^{(j)}
\end{aligned}
\end{equation}
where each $\gamma \sim Beta(a,a)$ for $a \in (0,\inf)$.




\section{Isotonic Data Augmentation}
\label{sec:isotonic}

In this section, we introduce the order restrictions in data augmentation for knowledge distillation, which is denoted as isotonic data augmentation. In Sec~\ref{sec:method:analysis}, we analyze the partial order restrictions of soft labels. We propose the new objective of knowledge distillation subjected to the partial order restrictions in Sec~\ref{sec:method:algorithm}. Since the partial order is a special directed tree, we propose a more efficient Adapted IRT algorithm based on IRT-BIN~\cite{pardalos1999algorithms} to calibrate the original soft labels. In Sec~\ref{sec:method:penalty}, we directly impose partial order restrictions on the student model. We propose a more efficient approximation objective based on penalty methods.

\subsection{Analysis of the Partial Order Restrictions}
\label{sec:method:analysis}
We hope that the soft label distribution by isotonic data augmentation and the hard label distribution have no order violations. We perform isotonic regression on the original soft labels $\mathcal{T}(\tilde{x})$ to obtain new soft labels that satisfy the order restrictions. We denote the new soft labels as the order restricted soft labels $m(\mathcal{T}(\tilde{x})) \in \mathbb{R}^c$. For simplicity, we will use $m$ to denote $m(\mathcal{T}(\tilde{x}))$. We use $m_i$ to denote the score of the $i$-th label.

To elaborate how we compute $m$, without loss of generality, we assume the indices of the two original labels of the augmented sample $(\tilde{x},\tilde{y})$ are $1,2$ respectively with $\gamma >0.5$. So $\tilde{y}$ is monotonically decreasing, i.e. $\tilde{y}_1 =\gamma > \tilde{y}_2 = 1-\gamma > \cdots > \tilde{y}_c$.

We consider two types of order restrictions: (1) the order between two original labels (i.e., $m_1 \ge m_2$); (2) The order between an original label and \cwy{a} non-original label (i.e. $\forall i\in\{1,2\},j\in\{3,\cdots,c\}, m_i \ge m_j$). For example, in Fig.~\ref{fig:intuition_draw}, we expect the probability of {\it panda} is greater than that of {\it cat}. And the probability of {\it cat} is greater than other labels. We do not consider the order between two non-original labels.

\begin{figure}[h]
\centering
\includegraphics[scale=0.45]{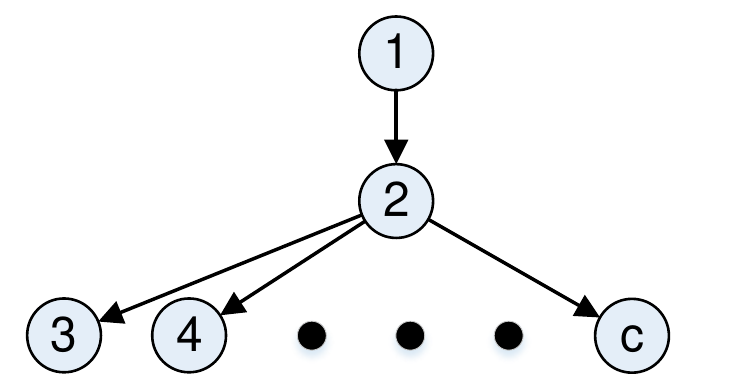}
\caption{The partial order restrictions is a directed tree.}
\label{fig:example_star} 
\vspace{-0.5cm}
\end{figure}

We use $G(V,E)$ to denote the partial order restrictions, where each vertex $i=1\cdots c$ represents $m_i$, an edge $(i,j) \in E$ represents the restriction of $m_i \ge m_j$. $E$ is formulated in Eq.~\eqref{eqn:partial}. We visualize the partial order restrictions in Fig.~\ref{fig:example_star}.
\begin{equation}
\label{eqn:partial}
\begin{aligned}
&E=\{(1,2)\} \cup \{(2,i)|i =3\cdots c\}
\end{aligned}
\end{equation}
\vspace{-0.5cm}

%

\begin{lemma}\label{lemma:tree}
$E$ is a directed tree.
\end{lemma}


\subsection{Knowledge Distillation via Order Restricted Soft Labels}
\label{sec:method:algorithm}
For an augmented sample $(\tilde{x},\tilde{y})$, we first use the teacher model to predict its soft labels. Then, we calibrate the soft labels to meet the order restrictions. We use the order-restricted soft label distribution $m$ to supervise the knowledge distillation. We formulate this process below.

{\bf Objective with Order Restricted Soft Labels.}
Given the hard label distribution $\tilde{y}$ and soft label distribution $\mathcal{T}(\tilde{x})$ of an augmented sample $(\tilde{x},\tilde{y})$, the objective of knowledge distillation with isotonic data augmentation is:
\begin{equation}
\label{eqn:pdef}
\begin{aligned}
&\loss_\text{KD-i} = \loss_\text{KD-aug} + \beta E_{(\tilde{x},\tilde{y}) \sim \mathcal{D}_{train}} CE(\tilde{y},\hat{m})
\end{aligned}
\end{equation}
where $\hat{m}$ denotes the optimal calibrated soft label distribution.

To compute $\hat{m}$, we calibrate the original soft label $\mathcal{T}(\tilde{x})$ to meet the order restrictions. There are multiple choices for $\hat{m}$ to meet the restrictions. Besides order restrictions, we also hope that the distance between the original soft label distribution $\mathcal{T}(\tilde{x})$ and the calibrated label distribution $m$ is minimized.
Intuitively, the original soft labels contain the knowledge of the teacher model. So we want this knowledge to be retained as much as possible. We formulate the calibration below.

We compute $\hat{m}$ which satisfies the order restriction $E$ while preserving most knowledge by minimizing the mean square error to the original soft labels:
\begin{subequations}
  \begin{equation}\label{eqn:isotonic:1}
    \hat{m} = \argmin_m mean\_square\_error(\mathcal{T}(\tilde{x}), m)
  \end{equation}
  \begin{equation}\label{eqn:isotonic:2}
    \text{subject to } \forall (i,j) \in E, \hat{m}_i \ge \hat{m}_j
  \end{equation}
\end{subequations}
Eq.~\eqref{eqn:isotonic:2} denotes the order restrictions. Eq.~\eqref{eqn:isotonic:1} denotes the objective of preserving most original information.
The goal of computing $\hat{m}$ can be reduced to the classical isotonic regression in statistics.

Here we analyze the difference between isotonic data augmentation and probability calibration in machine learning~\cite{niculescu2005predicting}. Isotonic regression is also applied in probability calibration. While both the proposed isotonic data augmentation and probability calibration try to rectify the erroneous predicted by models, our proposed isotonic data augmentation only happens in the training phase when the groundtruth distribution (i.e. the hard labels) is known. We use the isotonic soft labels $\hat{m}$ as the extra supervision for model training. In contrast, the probability calibration needs to learn an isotonic function and uses it to predict the probability of unlabeled samples.

{\bf Algorithm.} To optimize $\loss_\text{KD-i}$, we need to compute $\hat{m}$ first.
According to lemma~\ref{lemma:tree}, finding the optimal $\hat{m}$ can be reduced to the tree-structured IR problem, which can be solved by
IRT-BIN~\cite{pardalos1999algorithms} with binomial heap in $O(c\log c)$ time complexity. We notice that the tree structure in our problem is special: a star (nodes $2\cdots c$) and an extra edge $(1,2)$. So we give a more efficient implementation compared to IRT-BIN with only one sort in algorithm~\ref{algo}.


\begin{algorithm}[!htb]
\small
\caption{\bf Adapted IRT.}
\label{algo}
\hspace*{\algorithmicindent} \textbf{Data:} $\mathcal{T}(\tilde{x})$;
\begin{algorithmic}[1]
    \STATE Initialize $m_i \leftarrow \mathcal{T}(\tilde{x})_i$, $B_i \leftarrow \{i\}$ for $i=1\cdots c$
    \STATE Sort $m_i$ for $i=3\cdots c$ in descending order
    \STATE $i \leftarrow 3$
    \WHILE {$i\le c$ AND $m_2 < m_i$} \label{line:4}
        \STATE $m_2 \leftarrow \frac{m_2 \times |B_2|+m_i}{|B_2| +1}$ 
        \STATE $B_2 \leftarrow B_2 \cup \{i\}$
        \STATE $i \leftarrow i+1$ \label{line:7}
    \ENDWHILE

    \IF {$m_1 < m_2$} \label{line:8}
         \STATE $m_1 \leftarrow \frac{m_1+m_2 \times |B_2|}{|B_2|+1}  $
         \STATE $B_1 \leftarrow B_1 \cup B_2$ \label{line:10}
         \WHILE {$i\le c$ AND $m_1 < m_i$} \label{line:11}
            \STATE $m_1 \leftarrow \frac{m_1 \times |B_1|+m_i}{|B_1| +1}  $
            \STATE $B_1 \leftarrow B_1 \cup \{i\}$
            \STATE $i \leftarrow i+1$ \label{line:14}
        \ENDWHILE
    \ENDIF
    
\STATE Recover $\hat{m}$ from $m$ according to $B$
\STATE \textbf{Return} $\hat{m}$
\end{algorithmic}
\end{algorithm}

The core idea of the algorithm is to iteratively reduce the number of violations by merging node blocks until no order violation exists.  Specifically, we divide the nodes into several blocks, and use $B_i$ to denote the block for node $i$. At initialization, each $B_i$ only contains node $i$ itself. Since all nodes except $1$ and $2$ are leaf nodes with a common parent $2$, we first consider the violations between  $2$ and $i=3\cdots c$ (line~\ref{line:4}-\ref{line:7}). Note that nodes $i=3 \cdots c$ are sorted according to their soft probabilities $\mathcal{T}(\tilde{x})_i$. We enumerate $i=3 \cdots c$ and iteratively determine whether there is a violation between node $2$ and node $i$. If so, we absorb node $i$ into $B_2$. This absorption will set all nodes in $B_2$ to their average value. In this way, we ensure that there are no violations among nodes $2 \cdots c$. Then, we consider the order between $1$ and $2$. If they are discordant (i.e. $m_1<m_2$), we similarly absorb $B_2$ into $B_1$ to eliminate this violation (line~\ref{line:8}-\ref{line:10}). If this absorption causes further violations between $2$ and a leaf node, we similarly absorb the violated node as above (line~\ref{line:11}-\ref{line:14}). Finally, we recover $\hat{m}$ from $m$ according to the final block divisions.

\begin{theorem}\cite{pardalos1999algorithms}
The Adapted IRT algorithm terminates with the optimal solution to $\hat{m}$.
\end{theorem}

The correctness of the algorithm is due to the strictly convex function of isotonic data augmentation subject to convex constraints. Therefore it has a unique local minimizer which is also the global minimizer~\cite{bazaraa1979nonlinear}. Its time complexity is $O(c\log c)$.

\subsection{Efficient Approximation via Penalty Methods}
\label{sec:method:penalty}
We found two drawbacks of the proposed order restricted data augmentation in Sec~\ref{sec:method:algorithm}: (1) although the time complexity is $O(c\log c)$, the algorithm is hard to compute in parallel in GPU; (2) The order restrictions are too harsh, which overly distorts information of the original soft labels. For example, if the probability of original labels are very low, then almost all nodes will be absorbed and averaged. This will loss all valid knowledge from the original soft labels. In this subsection, we loose the order restrictions and propose a more GPU-friendly algorithm.

Note that, the partial order $E$ in Eq.~\eqref{eqn:isotonic:2} introduces the restrictions to the soft labels, and then uses the isotonic soft labels to limit the student model. If we directly use the partial order to limit the student model instead, the restrictions can be rewritten as:
\begin{equation}
\label{eqn:constraint_student}
\small
\begin{aligned}
& \forall (i,j) \in E, \mathcal{S}(\tilde{x})_i \ge \mathcal{S}(\tilde{x})_j \\
\Leftrightarrow & \mathcal{S}(\tilde{x})_1 \ge \mathcal{S}(\tilde{x})_2 \text{ AND } min(\mathcal{S}(\tilde{x})_{1,2}) \ge max(\mathcal{S}(\tilde{x})_{3 \cdots c})
\end{aligned}
\end{equation}
Note that we can replace $min(\mathcal{S}(\tilde{x})_{1,2}) \ge max(\mathcal{S}(\tilde{x})_{3 \cdots c})$ with a simpler term $\mathcal{S}(\tilde{x})_2 \ge max(\mathcal{S}(\tilde{x})_{3 \cdots c})$ without changing the actual restriction. We use $min(\mathcal{S}(\tilde{x})_{1,2}) \ge max(\mathcal{S}(\tilde{x})_{3 \cdots c})$ because we want to ensure the loss below is equally sensitive to both $\mathcal{S}(\tilde{x})_1$ and $\mathcal{S}(\tilde{x})_2$.

{\bf Objective with Order Restricted Student.} We convert the optimization problem subjected to Eq.~\eqref{eqn:constraint_student} to the unconstraint case in Eq.~\eqref{eqn:kdp} via penalty methods. The idea is to add the restrictions in the loss function.
\begin{equation}
\small
\label{eqn:kdp}
\begin{aligned}
&\loss_\text{KD-p}= \loss_\text{KD-aug} + \sigma E_{(\tilde{x},\tilde{y}) \sim \mathcal{D}_{train}} [max(0, \mathcal{S}(\tilde{x})_2 -\mathcal{S}(\tilde{x})_1 ) \\
& +  max(0, max(\mathcal{S}(\tilde{x})_3 \cdots \mathcal{S}(\tilde{x})_{c}) - min(\mathcal{S}(\tilde{x})_1, \mathcal{S}(\tilde{x})_2) )]
\end{aligned}
\end{equation}
where $\sigma$ is the penalty coefficients.
The penalty-based loss $\loss_\text{KD-p}$ can be computed in $O(c)$ time and is GPU-friendly (via the max function).

\section{Experiments}
\subsection{Setup}

\quad {\bf Models.} We use teacher models and the student models of different architectures to test the effect of our proposed isotonic data augmentation algorithms for knowledge distillation. We tested the knowledge transfer of the same architecture (e.g. from ResNet101 to ResNet18), and the knowledge transfer between different architectures (e.g. from GoogLeNet to ResNet).

{\bf Competitors.} We compare the isotonic data augmentation-based knowledge distillation with standard knowledge distillation~\cite{hinton2015distilling}. We also compare with the baseline of directly distilling with augmented samples without introducing the order restrictions. We use this baseline to verify the effectiveness of the order restrictions.

{\bf Datasets.} We use {\it CIFAR-100}~\cite{krizhevsky2009learning}, which contains 50k training images with 500 images per class and 10k test images. We also use ImageNet, which contains 1.2 million images from 1K classes for training and 50K for validation, to evaluate the scalability of our proposed algorithms.

{\bf Implementation Details.} For CIFAR-100, we train the teacher model for 200 epochs and select the model with the best accuracy on the validation set. The knowledge distillation is also trained for 200 epochs. We use SGD as the optimizer. We initialize the learning rate as 0.1, and decay it by 0.2 at epoch 60, 120, and 160. By default, we set $\beta=3, \sigma=2$, which are derived from grid search in $\{0.5,1,2,3,4,5\}$. We set $\tau=4.5, \alpha=0.95$ from common practice. For ImageNet, we train the student model for 100 epochs. We use SGD as the optimizer with initial learning rate is 0.1. We decay the learning rate by 0.1 at epoch 30, 60, 90. We also set $\beta=3, \sigma=2$. We follow~\cite{matsubara2020torchdistill} to set $\tau=1.0, \alpha=0.5$.
Models for ImageNet were trained on 4 Nvidia Tesla V100 GPUs. Models for CIFAR-100 were trained on a single Nvidia Tesla V100 GPU.

\subsection{Main Results}

\begin{table*}
\centering
\setlength{\tabcolsep}{2.1pt}
\small
\begin{tabular}{lccccccccc}
\hline
                        & ResNet101      & ResNet50       & ResNext50      & GoogleNet      & DenseNet121      & SeResNet101    & SeResNet101    & DenseNet121    &                   \\
                        & ResNet18       & ResNet18       & ResNet18       & ResNet18       & ResNet18       & ResNet18       & SeResNet18     & SeResNet18     & Avg.           \\ \hline \hline
Teacher                 & 78.28          & 78.85          & 78.98          & 78.31          & 78.84          & 78.08          & 78.08          & 78.84          &                   \\
Student                 & 77.55          & 77.55          & 77.55          & 77.55          & 77.55          & 77.55          & 77.21          & 77.21          &                   \\ \hline
KD                      & 79.78          & 79.41          & 79.88          & 79.33          & 79.84          & 79.41          & 77.45          & 79.65          & 79.34          \\
\hline
(KD Mixup)KD-aug           & 79.39          & 79.75          & 80.14          & 80.15          & 79.75          & 78.35          & 78.94          & 79.52          & 79.50          \\
(KD Mixup)KD-i      & 79.75          & 80.13          & 80.35          & 80.25          & \textbf{80.38} & 79.73          & 78.83          & 80.01          & 79.93          \\
(KD Mixup)KD-p  & \textbf{80.56} & \textbf{80.45} & \textbf{80.67} & \textbf{80.35} & 80.36          & \textbf{80.11} & \textbf{79.25} & \textbf{80.49} & \textbf{80.28}    \\ \hline
(KD CutMix)KD-aug               & 79.73          & 80.02          & 80.19          & 79.71          & 79.77          & 79.19          & 78.55          & 80.23          & 79.67          \\
(KD CutMix)KD-i     & \textbf{79.95} & 80.02          & \textbf{80.67} & \textbf{79.98} & \textbf{80.27} & 79.51          & 79.05          & 80.45          & 79.99           \\
(KD CutMix)KD-p & 79.93          & \textbf{80.51} & 80.34          & 79.96          & 79.98          & \textbf{79.57} & \textbf{79.13} & \textbf{80.83} & \textbf{80.03} \\ \hline 
CRD          & 79.76          & 79.75          & 79.59          & 79.74          & 79.74 & 79.22 & 79.35 & 79.86 &79.63 \\ \hline
(CRD Mixup)CRD-aug    & 79.52          & 79.38          & 80.03          & 79.92          & 80.05	&79.69	&79.41	&80.43&79.81    \\
(CRD Mixup)CRD-i  & \textbf{79.97} & \textbf{79.84} & \textbf{80.49} & 80.01          & 80.15 & 79.45 & 79.77 & 80.47&80.01 \\
(CRD Mixup)CRD-p   & 79.91          & 79.82          & 80.04          & \textbf{80.16} & \textbf{81.03} & \textbf{79.93} & \textbf{80.19} & \textbf{80.65} &\textbf{80.21}\\ \hline
(CRD CutMix)CRD-aug    & 79.77          & 79.63          & 79.96          & 80.13          & 80.18&	79.17&	79.49&	80.37   &79.84 \\
(CRD CutMix)CRD-i & \textbf{80.04} & 80.14          & \textbf{80.62} & \textbf{80.37} & \textbf{80.59} & 79.56 & 79.51 & \textbf{80.52} &\textbf{80.17}\\
(CRD CutMix)CRD-p & 79.91          & \textbf{80.19} & 80.11          & 80.28          & \textbf{80.59} & \textbf{79.77} & \textbf{80.01} & 80.48 &\textbf{80.17}\\ \hline
\end{tabular}
\caption{Results of CIFAR-100. KD means standard knowledge distillation~\protect\cite{hinton2015distilling} and CRD means contrastive representation distillation~\protect\cite{tian2019contrastive}. $*-$aug means knowledge distillation using mixup-based data augmentation without calibrating the soft labels, $*-i$ means soft labels by isotonic regression and $*-p$ means soft labels by the efficient approximation.}
\label{tab:result_cifar100}
\vspace{-0.5cm}
\end{table*}

{\bf Results on CIFAR-100.} We show the classification accuracies of the standard knowledge distillation and our proposed isotonic data augmentation in Table~\ref{tab:result_cifar100}. Our proposed algorithms effectively improve the accuracies compared to the standard knowledge distillation. This finding is applicable to different data augmentation techniques (i.e. CutMix and Mixup) and different network structures.
In particular, the accuracy of our algorithms even outperform the teacher models. This shows that by introducing the order restriction, our algorithms effectively calibrate the soft labels and reduce the error from the teacher model. As Mixup usually performs better than CutMix, we only use Mixup as data augmentation in the rest experiments.

{\bf Results on ImageNet.} We display the experimental results on ImageNet in Table~\ref{tab:result_imagenet}. We use the same settings as~\cite{tian2019contrastive}, namely using ResNet-34 as the teacher and ResNet-18 as the student. The results show that isotonic data augmentation algorithms are more effective than the original data augmentation technology. This validates the scalability of the isotonic data augmentation. 

We found that KD-p is better on CIFAR-100, while KD-i is better on ImageNet. We think this is because ImageNet has more categories (i.e. 1000), which makes order violations more likely to appear. Therefore, strict isotonic regression in KD-i is required to eliminate order violations. On the other hand, since CIFAR-100 has fewer categories, the original soft labels are more accurate. Therefore, introducing loose restrictions through KD-i is enough. As a result, we suggest to use KD-i if severe order violation occurs.


\begin{table}[!htb]
\centering
\setlength{\tabcolsep}{2.2pt}
\small
\begin{tabular}{lccc}
\hline
       & KD-aug & KD-i & KD-p \\
\hline \hline
top-1/top-5  & 68.79/88.24                              & {\bf 69.71/89.85}                           & 69.04/88.93\\ 
\hline
\end{tabular}
\caption{Results of ImageNet.}
\label{tab:result_imagenet}
\vspace{-0.3cm}
\end{table}

{\bf Ablation.} 
In Table~\ref{tab:result_cifar100}, we also compare with the conventional data augmentation without introducing order restrictions (i.e. KD-aug). It can be seen that by introducing the order restriction, our proposed isotonic data augmentation consistently outperforms the conventional data augmentation. This verifies the advantages of our isotonic data augmentation over the original data augmentation.

\subsection{Effect of Order Restrictions}
\label{sec:exp:order}
\begin{figure}[t]
\centering
\begin{minipage}[b]{0.225\textwidth}
	\centering
		\includegraphics[scale=.25]{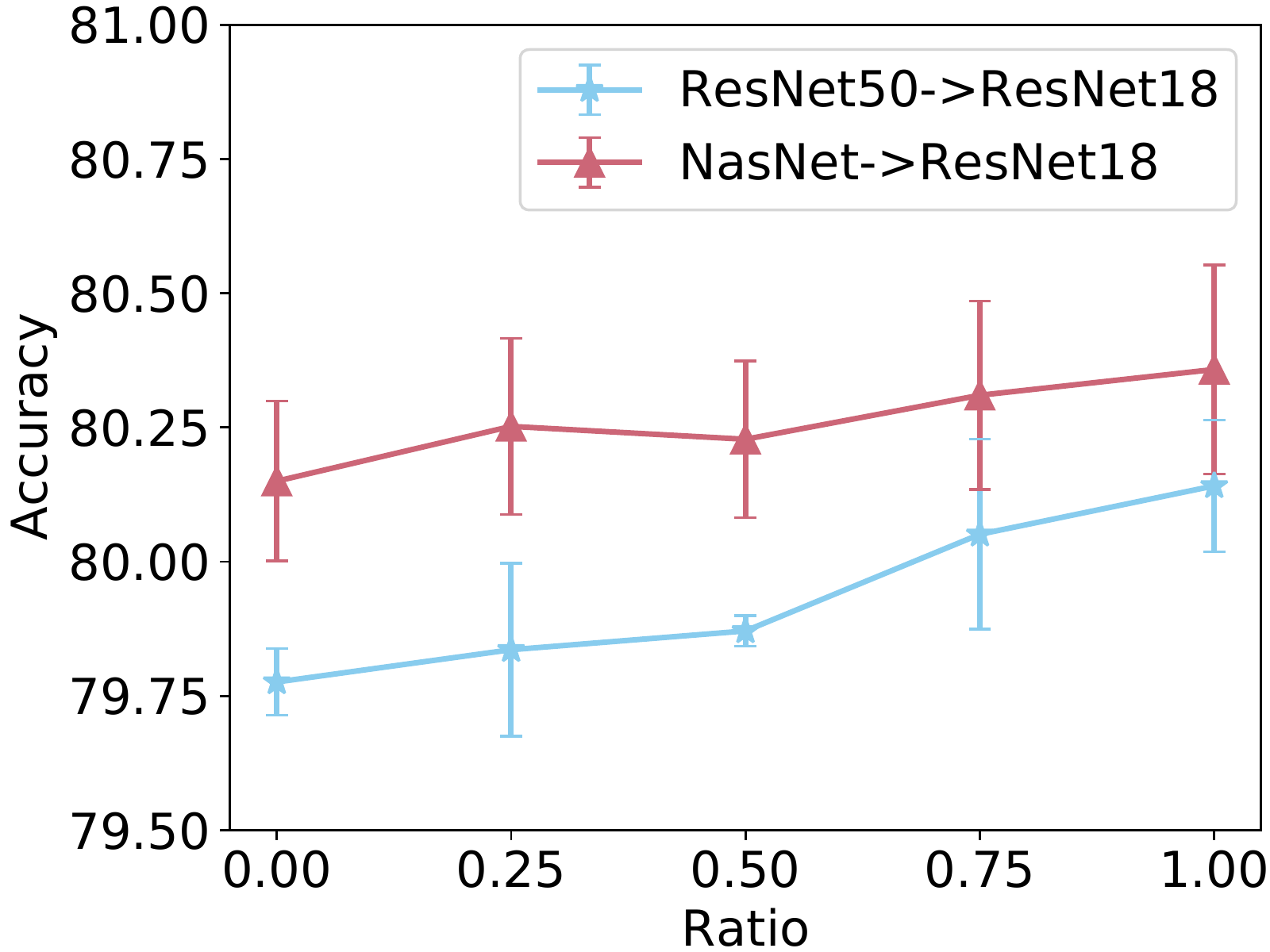}
\caption{Effect of introducing order restrictions to different ratios of samples. Average over 5 runs. Restricting more samples will improve the effect.}
\label{fig:isotonic_ratio}
\end{minipage}
\hspace{0.2cm}
\begin{minipage}[b]{0.225\textwidth}
	\centering
		\includegraphics[scale=.25]{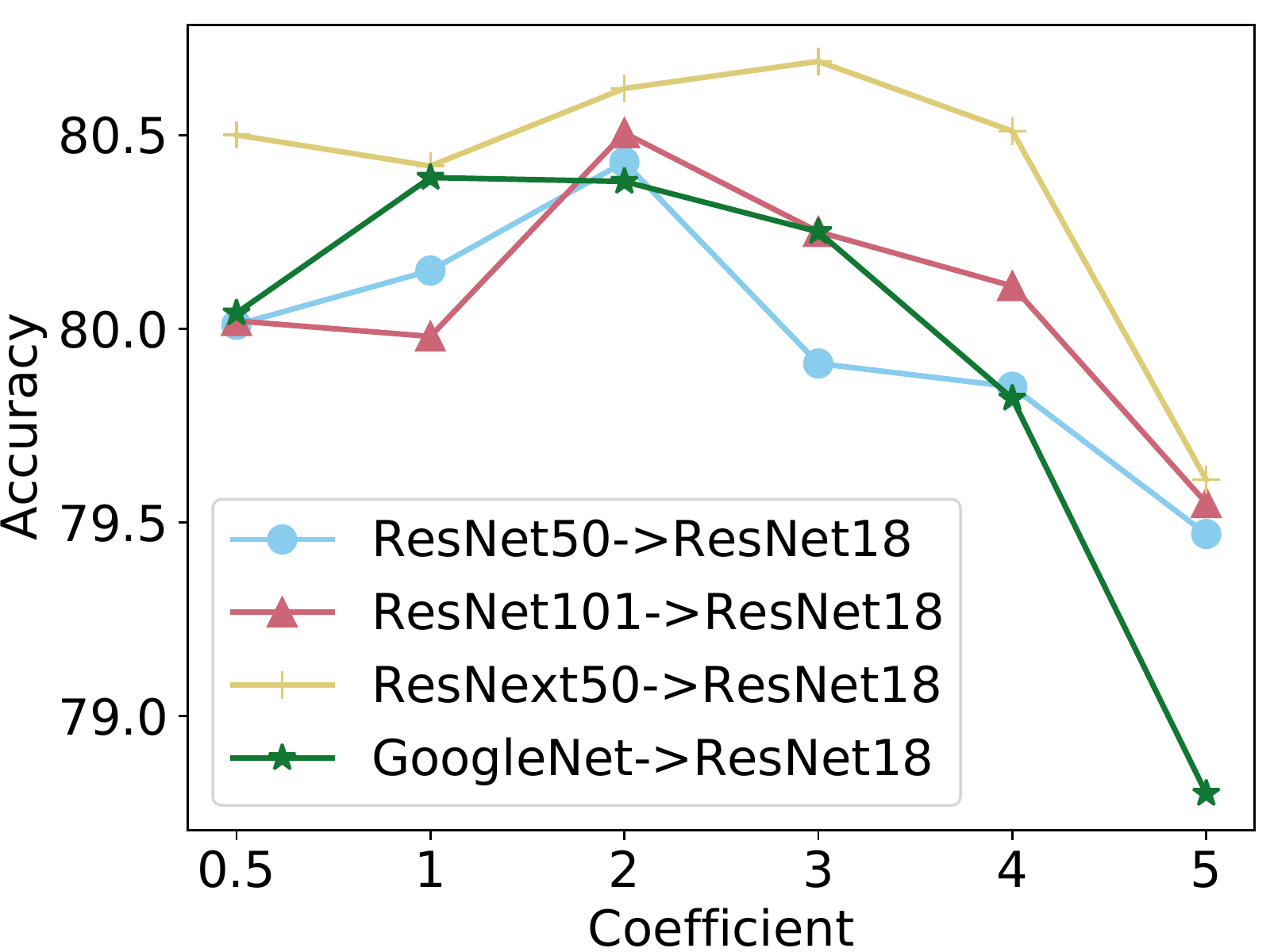}
\caption{Effect of different $\sigma$s. $\sigma=2$ is a recommended value as it outperforms other values in most cases. \newline}
	\label{fig:sigma}
\end{minipage}
\end{figure}

Our basic intuition of this paper is that, order violations of soft labels will injure the knowledge distillation. In order to verify this intuition more directly, we evaluated the performance of knowledge distillation under different levels of order violations. Specifically, we use the Adapted IRT algorithm to eliminate the order violations of soft labels for $0\%, 25\%, \cdots, 100\%$ augmented samples, respectively. We show in Fig.~\ref{fig:isotonic_ratio} the effectiveness of eliminating different proportions of order violations in CIFAR-100. As more violations are calibrated, the accuracy of knowledge distillation continues to increase. This verifies that the order violations injure the knowledge distillation.


%

\subsection{Efficiency of Isotonic Data Augmentation}
We mentioned that KD-p based on penalty methods is more efficient and GPU-friendly than KD-i. In this subsection, we verified the efficiency of different algorithms. We selected models from Table~\ref{tab:result_cifar100} and counted their average training time of one epoch. In Table~\ref{tab:time}, taking the time required for standard KD as the unit 1, we show the time of different data augmentation algorithms. It can be seen that KD-p based on penalty methods require almost no additional time. This shows that KD-p is more suitable for large scale data in terms of efficiency.

\begin{table}[!tb]
\centering
\small
\begin{tabular}{lcccc}
\hline
       & KD & KD-aug & KD-i   & KD-p   \\ \hline \hline
Mixup  & 1.00x & 1.02x  & 3.33x   & 1.02x \\
CutMix & 1.00x & 1.01x   & 3.05x & 1.01x   \\ \hline
\end{tabular}
\caption{Time costs for different data augmentation algorithms.}
\label{tab:time}
\vspace{-0.3cm}
\end{table}

\subsection{Effect of the Looseness of Order Restrictions}
The coefficient $\sigma$ in the Eq.~\eqref{eqn:kdp} is the key hyper-parameter that controls the looseness of KD-p. 
It can be found that for most models, the model performs best when $\sigma=2.0$. Therefore, $\sigma=2$ is a recommended value for real tasks.

%

\subsection{Effect on NLP Tasks}
Our proposed algorithm can also be extended to NLP tasks and Table \ref{tab:nlp} shows the results on several NLP tasks including SST \cite{socher2013recursive}, TREC \cite{li2002learning} and DBPedia\cite{auer2007dbpedia}. We use Bert\cite{devlin2019bert} as the teacher and DistilBert\cite{sanh2019distilbert} as the student. We leverage the mixup method in Mixup-Transformer\cite{sun2020mixuptransformer}, and the results indicate that comparing to KD-aug, KD-i and KD-p will improve student models' accuracy.

\begin{table}[!tb]
\centering
\small
\begin{tabular}{lccc}
\hline
                    & \multicolumn{1}{l}{SST} & \multicolumn{1}{l}{TREC} & \multicolumn{1}{l}{DBPedia} \\
\hline \hline
KD-aug & 97.35                   & 99.72                    & 98.54                      \\
KD-i     & 97.85                   & 99.78                    & 98.95                      \\
KD-p     & \textbf{98.24}                   & \textbf{99.95}                    & \textbf{99.01}                      \\
\hline
\end{tabular}
\caption{Results on several NLP tasks.}
\label{tab:nlp}
\vspace{-0.3cm}
\end{table}

\section{Conclusion}
We reveal that the conventional data augmentation techniques for knowledge distillation have critical order violations. In this paper, we use  isotonic regression (IR) - a classic statistical algorithm - to eliminate the rank violations. We adapt the traditional IRT-BIN algorithm to the adapted IRT algorithm to generate concordant soft labels for augmented samples. We further propose a GPU-friendly penalty-based algorithm. We have conducted a variety of experiments in different datasets with different data augmentation techniques and verified the effectiveness of our proposed isotonic data augmentation algorithms. We also directly verified the effect of introducing rank restrictions on data augmentation-based knowledge distillation. 

\section*{Acknowledgements}
This paper was supported by National Natural Science Foundation of China (No. 61906116), by Shanghai Sailing Program (No. 19YF1414700).

{
\bibliography{kd_mixup_ordinal}

\begin{thebibliography}{}

\bibitem[\protect\citeauthoryear{Acton and Bovik}{1998}]{acton1998nonlinear}
Scott~T Acton and Alan~C Bovik.
\newblock Nonlinear image estimation using piecewise and local image models.
\newblock {\em TIP}, 7(7):979--991, 1998.

\bibitem[\protect\citeauthoryear{Auer \bgroup \em et al.\egroup
  }{2007}]{auer2007dbpedia}
S{\"o}ren Auer, Christian Bizer, Georgi Kobilarov, Jens Lehmann, Richard
  Cyganiak, and Zachary Ives.
\newblock Dbpedia: A nucleus for a web of open data.
\newblock In {\em The semantic web}, pages 722--735. Springer, 2007.

\bibitem[\protect\citeauthoryear{Barlow and
  Brunk}{1972}]{barlow1972statistical}
Richard~E Barlow and Hugh~D Brunk.
\newblock The isotonic regression problem and its dual.
\newblock {\em JASA}, 67(337):140--147, 1972.

\bibitem[\protect\citeauthoryear{Bazaraa \bgroup \em et al.\egroup
  }{2013}]{bazaraa1979nonlinear}
Mokhtar~S Bazaraa, Hanif~D Sherali, and Chitharanjan~M Shetty.
\newblock {\em Nonlinear programming: theory and algorithms}.
\newblock John Wiley \& Sons, 2013.

\bibitem[\protect\citeauthoryear{Bochkovskiy \bgroup \em et al.\egroup
  }{2020}]{bochkovskiy2020yolov4}
Alexey Bochkovskiy, Chien-Yao Wang, and Hong-Yuan~Mark Liao.
\newblock Yolov4: Optimal speed and accuracy of object detection.
\newblock {\em arXiv preprint arXiv:2004.10934}, 2020.

\bibitem[\protect\citeauthoryear{Cho and Hariharan}{2019}]{cho2019efficacy}
Jang~Hyun Cho and Bharath Hariharan.
\newblock On the efficacy of knowledge distillation.
\newblock In {\em ICCV}, pages 4794--4802, 2019.

\bibitem[\protect\citeauthoryear{Das \bgroup \em et al.\egroup
  }{2020}]{das2020empirical}
Deepan Das, Haley Massa, Abhimanyu Kulkarni, and Theodoros Rekatsinas.
\newblock An empirical analysis of the impact of data augmentation on knowledge
  distillation.
\newblock {\em arXiv preprint arXiv:2006.03810}, 2020.

\bibitem[\protect\citeauthoryear{Devlin \bgroup \em et al.\egroup
  }{2019}]{devlin2019bert}
Jacob Devlin, Ming-Wei Chang, Kenton Lee, and Kristina Toutanova.
\newblock Bert: Pre-training of deep bidirectional transformers for language
  understanding, 2019.

\bibitem[\protect\citeauthoryear{Ding \bgroup \em et al.\egroup
  }{2019}]{ding2019adaptive}
Qianggang Ding, Sifan Wu, Hao Sun, Jiadong Guo, and Shu-Tao Xia.
\newblock Adaptive regularization of labels.
\newblock {\em arXiv preprint arXiv:1908.05474}, 2019.

\bibitem[\protect\citeauthoryear{Harris \bgroup \em et al.\egroup
  }{2020}]{harris2020fmix}
Ethan Harris, Antonia Marcu, Matthew Painter, Mahesan Niranjan, and Adam
  Pr{\"u}gel-Bennett~Jonathon Hare.
\newblock Fmix: Enhancing mixed sample data augmentation.
\newblock {\em arXiv preprint arXiv:2002.12047}, 2(3):4, 2020.

\bibitem[\protect\citeauthoryear{Hinton \bgroup \em et al.\egroup
  }{2015}]{hinton2015distilling}
Geoffrey Hinton, Oriol Vinyals, and Jeff Dean.
\newblock Distilling the knowledge in a neural network.
\newblock {\em arXiv preprint arXiv:1503.02531}, 2015.

\bibitem[\protect\citeauthoryear{Kendall}{1938}]{10.2307/2332226}
M.~G. Kendall.
\newblock A new measure of rank correlation.
\newblock {\em Biometrika}, 30(1/2):81--93, 1938.

\bibitem[\protect\citeauthoryear{Krizhevsky \bgroup \em et al.\egroup
  }{2009}]{krizhevsky2009learning}
Alex Krizhevsky, Geoffrey Hinton, et~al.
\newblock Learning multiple layers of features from tiny images.
\newblock 2009.

\bibitem[\protect\citeauthoryear{Li and Roth}{2002}]{li2002learning}
Xin Li and Dan Roth.
\newblock Learning question classifiers.
\newblock In {\em COLING}, 2002.

\bibitem[\protect\citeauthoryear{Mahsereci \bgroup \em et al.\egroup
  }{2017}]{mahsereci2017early}
Maren Mahsereci, Lukas Balles, Christoph Lassner, and Philipp Hennig.
\newblock Early stopping without a validation set.
\newblock {\em arXiv preprint arXiv:1703.09580}, 2017.

\bibitem[\protect\citeauthoryear{Matsubara}{2021}]{matsubara2020torchdistill}
Yoshitomo Matsubara.
\newblock torchdistill: A modular, configuration-driven framework for knowledge
  distillation.
\newblock In {\em International Workshop on Reproducible Research in Pattern
  Recognition}, pages 24--44, 2021.

\bibitem[\protect\citeauthoryear{Maxwell and
  Muckstadt}{1985}]{maxwell1985establishing}
William~L Maxwell and John~A Muckstadt.
\newblock Establishing consistent and realistic reorder intervals in
  production-distribution systems.
\newblock {\em OR}, 33(6):1316--1341, 1985.

\bibitem[\protect\citeauthoryear{Niculescu-Mizil and
  Caruana}{2005}]{niculescu2005predicting}
Alexandru Niculescu-Mizil and Rich Caruana.
\newblock Predicting good probabilities with supervised learning.
\newblock In {\em ICML}, pages 625--632, 2005.

\bibitem[\protect\citeauthoryear{Pardalos and
  Xue}{1999}]{pardalos1999algorithms}
Panos~M Pardalos and Guoliang Xue.
\newblock Algorithms for a class of isotonic regression problems.
\newblock {\em Algorithmica}, 23(3):211--222, 1999.

\bibitem[\protect\citeauthoryear{Sanh \bgroup \em et al.\egroup
  }{2019}]{sanh2019distilbert}
Victor Sanh, Lysandre Debut, Julien Chaumond, and Thomas Wolf.
\newblock Distilbert, a distilled version of bert: smaller, faster, cheaper and
  lighter.
\newblock {\em arXiv preprint arXiv:1910.01108}, 2019.

\bibitem[\protect\citeauthoryear{Socher \bgroup \em et al.\egroup
  }{2013}]{socher2013recursive}
Richard Socher, Alex Perelygin, Jean Wu, Jason Chuang, Christopher~D Manning,
  Andrew~Y Ng, and Christopher Potts.
\newblock Recursive deep models for semantic compositionality over a sentiment
  treebank.
\newblock In {\em EMNLP}, pages 1631--1642, 2013.

\bibitem[\protect\citeauthoryear{Sun \bgroup \em et al.\egroup
  }{2020}]{sun2020mixuptransformer}
Lichao Sun, Congying Xia, Wenpeng Yin, Tingting Liang, Philip~S Yu, and Lifang
  He.
\newblock Mixup-transfomer: Dynamic data augmentation for nlp tasks.
\newblock {\em arXiv preprint arXiv:2010.02394}, 2020.

\bibitem[\protect\citeauthoryear{Takahashi \bgroup \em et al.\egroup
  }{2019}]{takahashi2019data}
Ryo Takahashi, Takashi Matsubara, and Kuniaki Uehara.
\newblock Data augmentation using random image cropping and patching for deep
  cnns.
\newblock {\em TCSVT}, 2019.

\bibitem[\protect\citeauthoryear{Tian \bgroup \em et al.\egroup
  }{2019}]{tian2019contrastive}
Yonglong Tian, Dilip Krishnan, and Phillip Isola.
\newblock Contrastive representation distillation.
\newblock In {\em ICLR}, 2019.

\bibitem[\protect\citeauthoryear{Wang \bgroup \em et al.\egroup
  }{2020a}]{wang2020neural}
Dongdong Wang, Yandong Li, Liqiang Wang, and Boqing Gong.
\newblock Neural networks are more productive teachers than human raters:
  Active mixup for data-efficient knowledge distillation from a blackbox model.
\newblock In {\em CVPR}, pages 1498--1507, 2020.

\bibitem[\protect\citeauthoryear{Wang \bgroup \em et al.\egroup
  }{2020b}]{wang2020knowledgethrives}
Huan Wang, Suhas Lohit, Michael Jones, and Yun Fu.
\newblock Knowledge distillation thrives on data augmentation.
\newblock {\em arXiv preprint arXiv:2012.02909}, 2020.

\bibitem[\protect\citeauthoryear{Wen \bgroup \em et al.\egroup
  }{2019}]{wen2019preparing}
Tiancheng Wen, Shenqi Lai, and Xueming Qian.
\newblock Preparing lessons: Improve knowledge distillation with better
  supervision.
\newblock {\em arXiv preprint arXiv:1911.07471}, 2019.

\bibitem[\protect\citeauthoryear{Yang \bgroup \em et al.\egroup
  }{2019}]{yang2019training}
Chenglin Yang, Lingxi Xie, Siyuan Qiao, and Alan~L Yuille.
\newblock Training deep neural networks in generations: A more tolerant teacher
  educates better students.
\newblock In {\em AAAI}, volume~33, pages 5628--5635, 2019.

\bibitem[\protect\citeauthoryear{Yun \bgroup \em et al.\egroup
  }{2019}]{yun2019cutmix}
Sangdoo Yun, Dongyoon Han, Seong~Joon Oh, Sanghyuk Chun, Junsuk Choe, and
  Youngjoon Yoo.
\newblock Cutmix: Regularization strategy to train strong classifiers with
  localizable features.
\newblock In {\em ICCV}, pages 6023--6032, 2019.

\bibitem[\protect\citeauthoryear{Zhang \bgroup \em et al.\egroup
  }{2018}]{zhang2018mixup}
Hongyi Zhang, Moustapha Cisse, Yann~N Dauphin, and David Lopez-Paz.
\newblock mixup: Beyond empirical risk minimization.
\newblock In {\em ICLR}, 2018.

\end{thebibliography}
\bibliographystyle{named}
}

\end{document}